\documentclass{article}

\usepackage[preprint,nonatbib]{mainsty}

\usepackage[utf8]{inputenc} 
\usepackage[T1]{fontenc}    
\usepackage{hyperref}       
\usepackage{url}            
\usepackage{booktabs}       
\usepackage{amsfonts}       
\usepackage{nicefrac}       
\usepackage{microtype}      
\usepackage{xcolor}         
\usepackage{amsmath}
\usepackage{wrapfig}
\usepackage{pifont}
\usepackage{multirow}
\usepackage{enumitem}
\usepackage{algorithm}
\usepackage{algorithmic}
\usepackage[pdftex]{graphicx}
\title{AdaptMol: Adaptive Fusion from Sequence String to Topological Structure for Few-shot Drug Discovery}\usepackage[utf8]{inputenc} 

\author{
Yifan Dai$^1$\thanks{Yifan Dai and Xuanbai Ren have contributed equally to this work.}  , 
Xuanbai Ren$^1$$^*$, 
Tengfei Ma$^1$, 
Qipeng Yan$^2$, \\ 
\textbf{Yiping Liu$^1$,} 
\textbf{Yuansheng Liu$^1$,} 
\textbf{Xiangxiang Zeng$^1$\thanks{Corresponding author}}\\
$^1$College of Computer Science and Electronic Engineering, Hunan University\\
$^2$School of Biomedical Science, Hunan University\\
\{thintime, ren147, tfma, 	yuanshengliu, xzeng\}@hnu.edu.cn
}

\begin{document}

\maketitle

\begin{abstract}
Accurate molecular property prediction (MPP) is a critical step in modern drug development. However, the scarcity of experimental validation data poses a significant challenge to AI-driven research paradigms. Under few-shot learning scenarios, the quality of molecular representations directly dictates the theoretical upper limit of model performance. We present \textbf{AdaptMol}, a prototypical network integrating \textbf{\underline{Adapt}}ive multimodal fusion for \textbf{\underline{Mol}}ecular representation. This framework employs a dual-level attention mechanism to dynamically integrate global and local molecular features derived from two modalities: SMILES sequences and molecular graphs. (1) At the local level, structural features such as atomic interactions and substructures are extracted from molecular graphs, emphasizing fine-grained topological information; (2) At the global level, the SMILES sequence provides a holistic representation of the molecule.  To validate the necessity of multimodal adaptive fusion, we propose an interpretable approach based on identifying molecular active substructures to demonstrate that multimodal adaptive fusion can efficiently represent molecules. Extensive experiments on three commonly used benchmarks under 5-shot and 10-shot settings demonstrate that AdaptMol achieves state-of-the-art performance in most cases. The rationale-extracted method guides the fusion of two modalities and highlights the importance of both modalities.
\end{abstract}

\section{Introduction}
Drug discovery is essential for advancing public health and improving human well-being \cite{hu2020strategies,zeng2022accurate,zhang2023dualview}. However, the development of effective therapeutics currently demands substantial time and financial investment. In the past, researchers typically identified a large set of candidate molecules and conducted virtual screening to exclude those unlikely to exhibit the desired properties, thereby optimizing resource allocation and reducing potential waste \cite{riniker2013,sliwoski2014}. Recently, with the rapid advancement of artificial intelligence, deep learning models are increasingly being utilized for molecular property prediction. \cite{song2020communicative,fang2022molecular,fang2023knowledge}. However, many methods heavily rely on large quantities of labeled data, limiting their applicability in real-world scenarios where labeled data is scarce \cite{bertinetto2016learning}.

Few-shot learning has emerged as a transformative paradigm to address data scarcity in drug discovery, enabling models to generalize across novel molecular tasks with minimal samples. While graph neural networks (GNNs) naturally align with molecular graph structures by modeling atomic adjacencies and bond types \cite{altae-tran2017low,lv2024meta}, their effectiveness in low-data regimes is fundamentally constrained by three limitations: (1) overdependence on structural diversity in training data, (2) susceptibility to overfitting, and (3) compromised generalization to unseen molecular scaffolds. Recent multimodal approaches integrating GNNs with molecular fingerprints or SMILES partially enhance representation capacity \cite{bbae394,cai2022fp_gnn}, yet critical challenges persist: unaddressed inter-modal redundancy induces feature sparsity through vectors concatenation, while insufficient cross-modal interaction modeling fails to establish chemically meaningful relationships between descriptors. These deficiencies ultimately undermine the model's ability to distill pharmacologically relevant patterns, highlighting the urgent need for adaptive fusion mechanisms that balance information complementarity with redundancy mitigation while preserving domain-specific chemical insights.

To address the limitations of current molecular representation approaches in few-shot learning scenarios, we propose \textbf{Adaptive Fusion Prototype Networks for Molecules (AdaptMol)}. AdaptMol introduces a novel Adaptive Multi-level Attention (AMA) module, designed to extract and integrate molecular features from both local and global perspectives across multiple modalities. Specifically, AMA module dynamically fuses graph-based structural and topological information with high-dimensional SMILES representations derived from a large language model, assigning higher attention weights to the more informative modality on the  requirements of representation at either the local or global representation level. This adaptive weighting not only enhances the model’s ability to highlight the most salient molecular features but also effectively suppresses redundant or noisy information arising from modality misalignment. Importantly, we employ an interpretability-driven approach to assess the importance of dynamically fused multimodality and to improve the interpretability of the model inference process. Moreover, this approach also helps to identify key substructures that determine molecular activity, leading to more efficient exploration of the chemical space and discovery of novel effective drugs. Briefly, our contributions are summarized as follows: 

\begin{itemize}
\item We propose AdaptMol, a novel few-shot learning framework tailored for drug discovery tasks, capable of learning rich and generalizable molecular representations.
\item An adaptive fusion mechanism (AMA) is introduced to dynamically balance and align the two modalities, enabling multi-perspective learning of both structural and semantic features.
\item We propose a novel methodology to facilitate the identification of key substructures that influence molecular properties, thereby improving the explanation and interpretability of the model, and enhancing its overall credibility.
\item The AdaptMol tackles the issue of limited sample availability in drug discovery, offering a robust solution to the few-shot problem commonly encountered in this domain. Furthermore, experiments on three benchmarks show that AdaptMol can achieve state-of-the-art performance in most cases. We also conducted experiments on datasets from different domains to demonstrate the strong generalization capability of AdaptMol.
\end{itemize}

\section{Related work}
\textbf{Molecular multimodality learning.} The integration of information from modalities, such as SMILES, graph and molecular fingerprints, holds substantial potential for enhancing molecular representation. Graph modality effectively provides topological structures of molecules, while SMILES and molecular fingerprints encapsulate chemical semantics. However, most existing methods, including SMICLR \cite{pinheiro2022smiclr}, MOCO \cite{zhu2022improving}, and APN \cite{Hou2024}, adopt relatively naive strategies for multimodal fusion. Such approaches overlook the fact that simplistic concatenation can exacerbate feature sparsity and hinder the model's ability to capture meaningful cross-modal interactions. Our model, AdaptMol, introduces an adaptive multi-level attention module designed to enable more effective cross-modal interaction and information fusion, thereby improving the overall performance of molecular representation.\\ 
\textbf{Few-shot learning for molecular property prediction.} Few-shot learning (FSL) \cite{vinyals2016matching,chen2019closer} addresses scenarios with limited labeled data. Currently, drug discovery tasks face the challenge of data scarcity due to the difficulty in collecting, preprocessing and labeling data, so FSL has become a promising solution \cite{bansal2022systematic}. In recent years, an increasing number of FSL algorithms have adopted meta-learning strategies, which learn prior knowledge or task-specific experience from a distribution of related tasks, enabling rapid adaptation to new tasks with limited labeled data \cite{liu2020adaptive,yao2021meta,hospedales2021meta}. Meta-learning-based FSL methods are primarily categorized into two approaches: optimization-based and metric-based. Optimization-based methods, such as MAML \cite{finn2017model}, aim to learn model parameters that can be rapidly fine-tuned to new tasks using a few gradient steps. In contrast, metric-based methods, including Prototypical Networks \cite{snell2017prototypical}, focus on learning embeddings and similarity measures to classify new instances based on their proximity to labeled examples in the embedding space. In the field of molecular property prediction, optimization-based approaches have been extensively applied \cite{lv2024meta,wang2021property,finn2017model,guo2021few}. Conversely, metric-based methods remain underexplored and present promising avenues for future research \cite{altae2017low,vella2022few}. \\
\textbf{Explain model predictions using rationales.} Molecular representation learning, including GNNs, often relies on black-box models that lack interpretability \cite{rudin2019interpretable}. Prior work on interpretability has primarily focused on image and text classification domains \cite{ribeiro2016why,sundararajan2017axiomatic}. To bridge the gap, we transfer the interpretability techniques from these areas to moleculer property prediction. Specifically, we leverage our model's prediction scores to extract rationales that determine molecular activity \cite{jin2020multi}.

\begin{figure*}[t!]
    \centering
    \includegraphics[width=\textwidth]{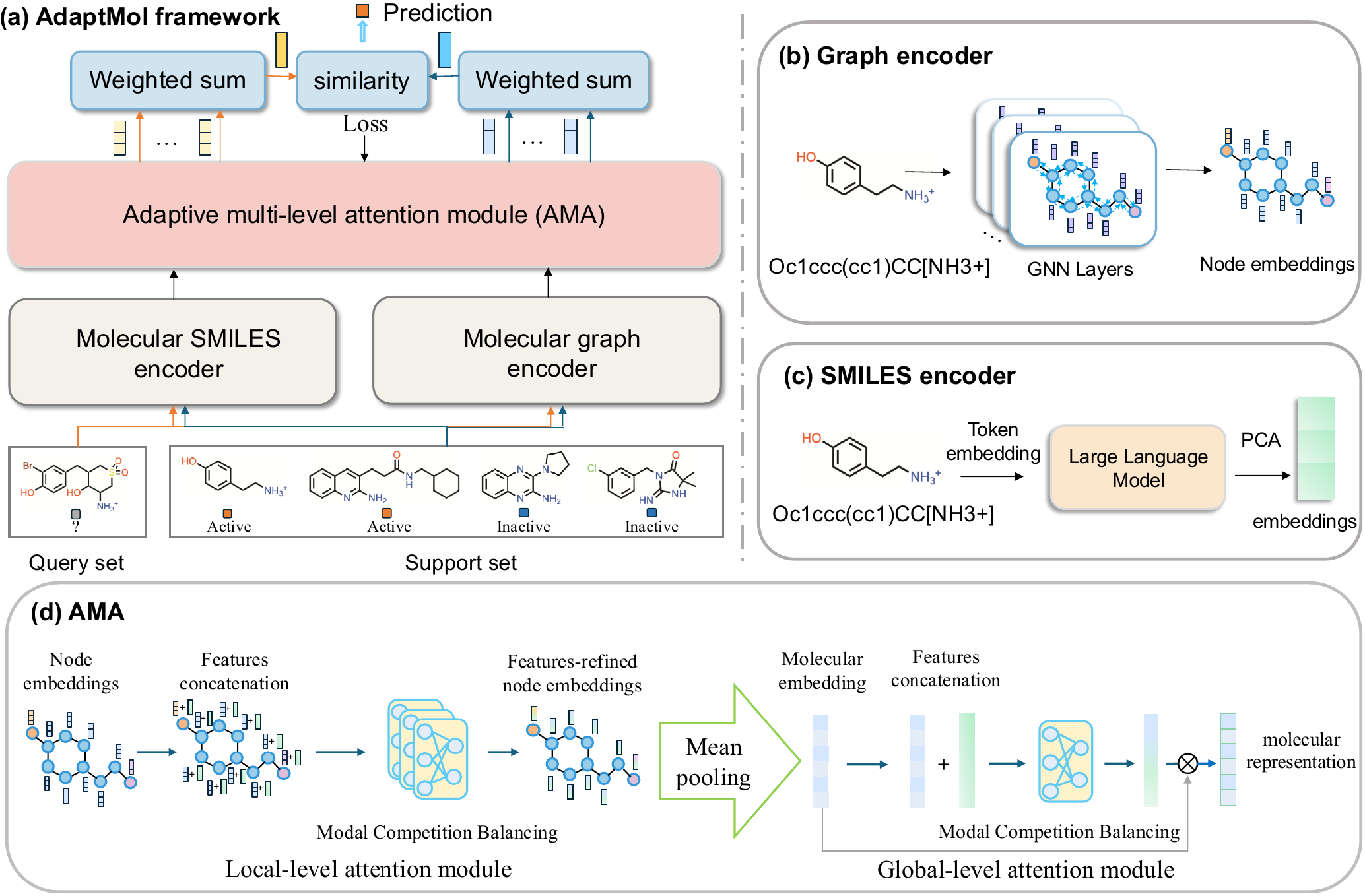} 
    \caption{(a) Overview of the proposed AdaptMol framework, where we plot a 2-way 2-shot task. AdaptMol is optimized over training tasks. Within each task $T_t$, the support set obtains prototypes for each class, while the query set optimizes the two molecular encoders and AMA module. During the testing phase, molecule in the query set is represented by the encoders and AMA module, used to compute similarity with the prototypes, leading to the final prediction. (b) The molecular graph encoder, generating molecular representations from the molecular graph. (c) Molecular SMILES encoder, using a large language model to capture the semantic and contextual information of molecular sequences. (d) The overall framework of the proposed AMA. The representation of all nodes within a molecule is sequentially processed through adaptive attention modules from local and global level, resulting in the final features-refined molecular representation.}
    \label{fig:AdaptMol_framework}
\end{figure*}

\section{Preliminaries}
This section introduces our proposed Adaptive Fusion Prototype Networks (AdaptMol). We begin by formulating the few-shot molecular prediction problem \ref{subsec1}. An overview of the methodologies employed follows \ref{subsec2}. Subsequently, we elaborate on the specifics of the encoders and describe the adaptive multi-level attention module that integrates sequence  syntactic features with graph topological information within AdaptMol \ref{subsec3}. Next, we focus on the application of generative models to elucidate the molecular rationale underlying predictions, emphasizing their role in enhancing the interpretability of the inferences drawn from the predictive model \ref{subsec4}. Finally, we elaborate the training and evaluation processes utilized in AdaptMol \ref{subsec5}.

\subsection{Problem Definition}\label{subsec1}

Following \cite{altae2017low,guo2021few}, few-shot molecular property prediction is conducted across a series of tasks $\mathcal{T}$, where each task $T$ involves predicting a specific molecular property $p$. The training set $\mathcal{L}_{\text{train}}$ consists of a set of tasks $\mathcal{T}_{\text{train}}$ and is represented as $\mathcal{L}_{\text{train}} = \{(x_i, y_i, t) | t \in \mathcal{T}_{\text{train}}\}$, where $x_i$ denotes the $i$-th molecule, $y_i$ denotes the label (property) of this molecule for task $t$. The test set $\mathcal{L}_{\text{test}}$ comprises a completely distinct set of tasks $\mathcal{T}_{\text{test}}$ and is expressed as $\mathcal{L}_{\text{test}} = \{(x_j, y_j, t) | t \in \mathcal{T}_{\text{test}}\}$. The property sets for the training and test tasks are denoted as $\mathcal{P}_{\text{train}}$ and $\mathcal{P}_{\text{test}}$, satisfying the condition: $\mathcal{P}_{\text{train}} \cap \mathcal{P}_{\text{test}} = \emptyset$. The objective of AdaptMol is to train on the training set $\mathcal{L}_{\text{train}}$, thereby learning a predictor capable of inferring novel molecular properties in the test set $\mathcal{L}_{\text{test}}$, where only a limited labeled molecules are available.

To address the few-shot problem, the episodic training paradigm in meta-learning has demonstrated remarkable effectiveness \cite{finn2017model}. During the training phase, we iteratively sample batches of episodes $\{\mathcal{E}_t\}_{t=1}^N$, where $N$ denotes the number of episodes, rather than loading the entire training dataset into memory. To construct an episode $\mathcal{E}_t$, we first sample a target task $T_t$ from the training tasks $\mathcal{T}_{\text{train}}$, followed by sampling a labeled support set $\mathcal{S}_t$ and an unlabeled query set $\mathcal{Q}_t$. In this case, we employ a 2-way $K$-shot episode, meaning that the support set $\mathcal{S}_t$ consists of two classes (i.e., active $y=1$ or inactive $y=0$), with $K$ molecules in each class, i.e., $\mathcal{S}_t = \{(x_i^s, y_i^s, t_i^s)\}_{i=1}^{2K}$, and query set $\mathcal{Q}_t = \{(x_i^q, y_i^q, t_i^q)\}_{i=1}^M$, where $M$ denotes the number of molecules in the query set. Finally, we define the episode as $\mathcal{E}_t = \{\mathcal{S}_t, \mathcal{Q}_t\}$.

\subsection{Overview of the method}\label{subsec2}
The overall architecture of the Adaptive Fusion Prototype Networks, as depicted in Figure \ref{fig:AdaptMol_framework} (a), consists primarily of two encoders and an adaptive multi-level attention module. The process begins with the application of a graph encoder, such as GIN, to generate molecular representations from the molecular graph. Subsequently, a secondary encoder captures molecular  syntactic features from the corresponding SMILES. In particular, the refinement step employs an adaptive multi-level attention module to integrate and interact with the sequence features derived from SMILES and the graph representation obtained by Graph encoder. This approach enhances their capacity to capture comprehensive and nuanced molecular representations. Finally, considering that each molecular representation within the support set contributes differently to the prototype, we computed the prototypes for positive and negative samples separately in a weighted manner.

\subsection{Encoders and AMA module}\label{subsec3}
For graph encoder, illustrated in \ref{fig:AdaptMol_framework} (b), all node representations are captured by GIN, denoted as

\begin{equation}\label{equa1}
G = \{ g_j \}_{j=1}^N \in \mathbb{R}^{N \times d^g},
\end{equation}

where $d^g$ represents the length of the node representations, and $N$ denotes the number of nodes.

As shown in Figure \ref{fig:AdaptMol_framework} (c), we transform masked SMILES tokens $T^M_S$ into token ids $ID^M_S$ and expand the vocabulary. We then apply a large language model to derive global features $F_S \in \mathbb{R}^{d}$, where $d$ is the feature dimension. To manage the high dimensionality, we apply Principal Component Analysis (PCA) to reduce the feature space to $d^a$ dimensions, resulting in the final global features as

\begin{equation}
a = \phi(F_S) \in \mathbb{R}^{d^a}.
\end{equation}

To better integrate the sequence features with graph representations, we propose an adaptive multi-level attention module (AMA). The detailed structure of this module is depicted in Figure \ref{fig:AdaptMol_framework} (d). The AMA module consists of a local-level attention module and a global-level attention module, which collaboratively guide the model to focus on critical molecular information across multiple levels. Considering the dominant modality varies across different levels, we introduce an adaptive weight $\beta$. It can be formulated as follows:

\begin{equation}
\label{betas}
\beta(g) =
\begin{cases}
\beta_{\text{min}} + (\beta_{\text{max}} - \beta_{\text{min}})
 \times k, & \text{if local level } , \\
 
\beta_{\text{mid}} - (\beta_{\text{mid}} - \beta_{\text{min}}) \times k, & \text{if global level},
\end{cases}
\end{equation}

\begin{equation}
\label{betag}
\beta(s) =
\begin{cases}
\beta_{\text{mid}} - (\beta_{\text{mid}} - \beta_{\text{min}}) \times k, & \text{if local level } , \\
\beta_{\text{min}} + (\beta_{\text{max}} - \beta_{\text{min}}) \times k, & \text{if global level},
\end{cases}
\end{equation}

where k is the scaling factor. $\beta_\text{min}$ and $\beta_\text{max}$ are the predefined minimum and maximum values of the weight, with $\beta_\text{mid}$ = ($\beta_\text{min}$ + $\beta_\text{max}$)/2. Then we define AMA's input as:
\begin{equation}
F_{\text{l\_input}} = \big[ g_j \cdot \beta(g) ; a \cdot \beta(s) \big]_{j=1}^N \in \mathbb{R}^{N \times (d^g + d^a)},
\end{equation}

where $[* ; *]$ denotes concatenation. Thereafter, a multi-head self-attention layer with a sigmoid activation function is employed to compute the local attention, where $\sigma$ denotes the sigmoid activation function:

\begin{equation}
\text{Attn}_{\text{local}} = \sigma\big(\text{MultiHead}(F_{\text{l\_input}}, F_{\text{l\_input}}, F_{\text{l\_input}})\big) \in \mathbb{R}^{N \times d^g}.
\end{equation}

To obtain node representations refined at the local level, we multiply $\text{Attn}_{\text{local}}$ with the node representations $G_i$, denoted as:
\begin{equation}
F_{\text{l\_output}} = \text{Attn}_{\text{local}} \otimes G_i = \{ g_j' \}_{j=1}^N \in \mathbb{R}^{N \times d^g},
\end{equation}
where $F_{\text{l\_output}} \in \mathbb{R}^{d^g}$ represents the output of the local-level attention module and $\otimes$ denotes the element-wise multiplication.

For the global-level attention module, we begin by calculating the average representation of all nodes for each molecule \( x_i \), \( g_i = \frac{1}{N} \sum_j g_j' \in \mathbb{R}^{d^g} \). Then the input to the module is denoted as \( F_{\text{g\_input}} = \big[ g_i \cdot \beta(g) ; a \cdot \beta(s) \big] \in \mathbb{R}^{(d^g + d^a)} \). We employ a fully connected layer followed by a sigmoid function to compute the global attention:

\begin{equation}
\text{Attn}_{\text{global}} = \sigma \big( f_{\text{global}}(F_{\text{g\_input}}) \big) \in \mathbb{R}^{d^g},
\end{equation}

where \( f_{\text{global}} \) denotes the fully connected layer. Finally, to obtain the final molecular representations, we multiply \( \text{Attn}_{\text{global}} \) with the node representations \( g_i \), denoted as

\begin{equation}\label{equa9}
F_{\text{g\_output}} = \text{Attn}_{\text{global}} \otimes g \in \mathbb{R}^{d^g},
\end{equation}

where \( F_{\text{g\_output}} \in \mathbb{R}^{d^g} \) is the final molecular representation refined through multimodal features at both the local and global levels.

\subsection{Deriving Molecular Rationales through Predictive Models}\label{subsec4}
A rationale $S^i$ for property $i$ is defined as a subgraph of molecule $G$ that satisfies:  

\begin{enumerate}
    \item $|S^i| \leq N_s = 20$ (small size).
    \item $r_i(S^i) \geq \delta_i$ (high predicted score).
\end{enumerate}

To extract rationales, we use AdaptMol predictions on positive molecules $D_i^{\text{positive}}$. For each $G_i^{\text{positive}} \in D_i^{\text{positive}}$, subgraphs $S^i \subseteq G_i^{\text{positive}}$ are identified that satisfy:
\[
r_i(S^i) \geq \delta_i, \quad |S^i| \leq N_s, \quad \text{and } S^i \text{ is connected.}
\]

Due to the exponential subgraph space, we constrain $S^i$ to connected subgraphs, identified by iteratively removing non-essential bonds while retaining core properties. This is formulated as a search problem, solved using Monte Carlo Tree Search (MCTS) \cite{silver2017mastering}.  

In MCTS, the root represents the positive molecule $G_i^{\text{positive}}$, and each state $s$ corresponds to a subgraph obtained by selective bond removals. To ensure chemical validity, deletions are restricted to peripheral non-aromatic bonds or rings. Key metrics include:  

\begin{enumerate}
    \item $N(s, a)$ represents the visitation count of deleting $a$, used to balance exploration and exploitation during the search process.
    \item $W(s, a)$ denotes the total action value of the edge, indicating the likelihood that deleting $a$ will lead to the generation of an excellent rationale.
    \item $Q(s, a)$ represents the average action value, $Q(s, a) = W(s, a) / N(s, a)$.
    \item $R(s, a) = r_i(s')$ corresponds to the predicted property score of the new subgraph $s'$ obtained by deleting $a$ from $s$.
\end{enumerate}

Each MCTS iteration consists of:  

\begin{wrapfigure}[12]{r}{0.5\textwidth} 
    \centering
    \includegraphics[width=0.495\textwidth]{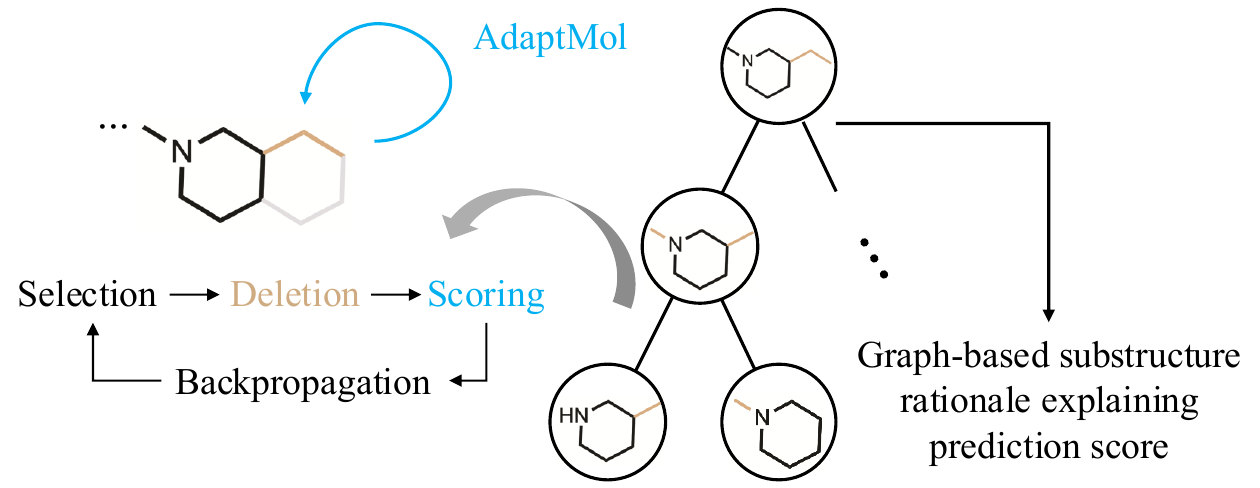} 
    \caption{Illustration of the Monte Carlo Tree Search (MCTS) method for deriving chemical structure rationales (graph substructures) associated with high predicted molecular activity.}
    \label{fig:intro}
\end{wrapfigure}

\noindent


\subsubsection*{1. Forward Propagation} \label{subsubsec1}
Select a path from the root $s_0$ to a leaf $s_L$ ($|s_L| \leq N_s$) and evaluate $r_i(s_L)$. At each state $s_k$, select the bond deletion $a_k$ as:  
\[
a_k = \arg \max_a Q(s_k, a) + U(s_k, a),
\]
\[
U(s_k, a) = c_{\text{puct}} R(s_k, a) \sqrt{\frac{\sum_b N(s_k, b)}{1 + N(s_k, a)}}.
\]
Here, $c_{\text{puct}}$ balances exploration and exploitation.  

\subsubsection*{2. Backward Propagation}  \label{subsubsec2}
Update statistics:  
\[
N(s_k, a_k) \leftarrow N(s_k, a_k) + 1
\]
\[
W(s_k, a_k) \leftarrow W(s_k, a_k) + r_i(s_L)
\]

Leaf nodes $s$ with $r_i(s) \geq \delta_i$ are added to the rationale vocabulary $V_S^i$. The detailed process is illustrated in Figure \ref{fig:intro}.

\subsection{Training and Evaluation}\label{subsec5}

AdaptMol is a model based on prototype networks, which implies that in a few-shot classification task, prototypes for each category must be computed. The refined molecular representations after the AMA model in a specific task are denoted as \(Z_t' = \{z_i'\}_{i=1}^{2K+M} \in \mathbb{R}^{d_g}\). The prototype representation of positive (negative) samples, \(p_{\text{positive}} (p_{\text{negative}})\), is calculated as a weighted sum of all positive (negative) samples. Specifically, for each embedded support point within a class, a distance is computed, representing the sum of Euclidean distances between the point and all other points, and the assigned weight is inversely proportional to this distance, meaning that larger distances result in smaller weight assignments. Formally, the positive prototype is calculated according to the following equation:

\begin{equation}\label{equa10}
\begin{cases}
p_{\text{positive}} = \sum_{i=1}^{K} \text{avg}_i z_i', & i \in [1, K] \\
\text{avg}_i = \frac{\text{weight}_i}{\sum_{j=1}^{K} \text{weight}_j}, & j \in [1, K] \\
\text{weight}_i = \frac{1}{\text{distance}_i} \\
\text{distance}_i = \sum_{j=1}^{K} \text{L2}(z_i', z_j'), & j \in [1, K]
\end{cases}
\end{equation}

The predicted labels of molecules in the query set are determined by the dot product similarity between AdaptMol-generated outputs for the molecules and the two prototypes. During the training phase, these predicted labels are used to compute the loss, which is subsequently utilized to update the model parameters:

\begin{equation}\label{equa11}
\begin{cases}
L_i = -[y_i \cdot \log(p_i) + (1 - y_i) \cdot \log(1 - p_i)], \\
\text{Loss}_t = \frac{1}{M} \sum_{i=1}^{M} L_i,  i \in [1, M]
\end{cases}
\end{equation}

In this context, \( y_i \) signifies the label of molecule \( i \), with 1 for positive and 0 for negative. The symbol \(p_i\) represents the predicted probability of molecule \(i\) being classified as a positive sample, which serves as the predicted label. During testing, predicted labels for the target task are used to characterize drug activity in corresponding molecules. The Appendix \ref{algori} provides Algorithm \ref{alg:algorithm}, detailing the AdaptMol training procedure.

\begin{table}[!t]
    \caption{ROC-AUC scores (\%) with standard deviations of all compared methods on MoleculeNet benchmark. The best results are highlighted in bold font.}
    \label{tab:performance_all}
    \centering
    \resizebox{\textwidth}{!}{ 
    \begin{tabular}{lcccccc}
        \toprule
         \multirow{2}{*}{Moldel} & \multicolumn{2}{c}{Tox21} & \multicolumn{2}{c}{SIDER} & \multicolumn{2}{c}{MUV} \\
        \cmidrule(lr){2-3} \cmidrule(lr){4-5} \cmidrule(lr){6-7} 
        & 5-shot & 10-shot & 5-shot & 10-shot & 5-shot & 10-shot \\
        \midrule
        Siamese & 63.34 (2.15) & 70.71 (1.40) & 52.69 (0.29) & 55.86 (0.93) & 49.94 (0.73) & 49.59 (0.86) \\
        AttnLSTM & 58.69 (1.69) & 65.97 (3.80) & 49.51 (0.84) & 49.18 (2.52) & 50.74 (0.49) & 50.99 (0.21) \\
        CHEF & 61.97 (0.65) & - & 57.34 (0.82) & - & 53.17 (4.21) & - \\
        ProtoNet & 72.78 (3.93) & 74.98 (0.32) & 64.09 (2.37) & 64.54 (0.89) & 58.31 (3.18) & 65.58 (3.11)\\
        MAML & 69.17 (1.34) & 80.21 (0.24) & 60.92 (0.65) & 70.43 (0.76) & 63.00 (0.61) & 63.90 (2.28)\\
        TPN & 75.45 (0.95) & 76.05 (0.24) & 66.52 (1.28) & 67.84 (0.95) & 65.13 (0.23) & 65.22 (5.82)\\
        BOIL & 76.75 (0.11) & 80.53 (0.20) & 67.97 (0.93) & 70.52 (0.42) & 60.13 (2.94) & 63.42 (2.09) \\ 
        EGNN & 76.80 (2.62) & 81.21 (0.16) & 60.61 (1.06) & 72.87 (0.73) & 63.46 (2.58) & 65.20 (2.08)\\
        IterRefLSTM & 75.09 (2.25) & 74.46 (0.21) & 66.52 (2.40) & 63.19 (2.23) & 50.95 (11.85) & 54.11 (13.82)\\
        PAR & 80.46 (0.13) & 82.06 (0.12) & 71.87 (0.48) & 
74.68(0.31) & 64.12(1.18) & 66.48(2.12) \\
        MetaGAT & 79.98 (0.11) & 82.40 (1.00) & 77.31 (0.20) & 77.73 (0.72) & 65.21(1.32) & 65.22(0.84) \\
        APN & 76.08 (0.23) & 78.02 (0.36) & 75.07 (0.38) & 79.02 (0.72) & 62.94 (0.66) & 63.69 (0.58) \\
        UniMatch & - & 82.62 (0.43) & - & 68.13 (1.54) & - & \textbf{79.40 (3.14)}\\
        AdaptMol & \textbf{83.79 (0.21)} & \textbf{84.93 (0.27)} & \textbf{79.60 (0.61)} & \textbf{81.59 (0.33)} & \textbf{71.65 (0.56)} & 77.16(0.54) \\
        \bottomrule
    \end{tabular}
    }
\end{table}

\section{Experiments}
\subsection{Experimental setting}

\begin{wraptable}[10]{r}{0.5\textwidth} 
  \caption{The detail information of datasets.}
  \label{tab:dataset_summary}
  \vspace{6pt}
  \centering
  \begin{tabular}{lccc}
    \toprule
    Datasets & Tox21 & SIDER & MUV \\
    \midrule
    Molecules & 7831 & 1427 & 93127 \\
    Tasks & 12 & 27 & 17 \\
    Training Tasks & 9 & 21 & 12 \\
    Testing Tasks & 3 & 6 & 5 \\
    \bottomrule
  \end{tabular}
\end{wraptable}

\textbf{Datasets and evaluation protocol.} Our study utilized three widely recognized datasets from MoleculeNet \cite{wu2018moleculenet} for few-shot molecular property prediction, and the data splitting strategy outlined in \cite{altae2017low} was subsequently employed. Table \ref{tab:dataset_summary} and Appendix \ref{dataset} provide a detailed summary of these datasets, including the number of molecules, the total number of tasks, and the division of tasks into training and testing subsets. During the evaluation phase, we followed the methodology outlined in \cite{wang2021property}, leveraging ROC-AUC as the evaluation metric to assess the performance of our proposed model in comparison to other baseline methods. We conducted ten independent experiments and reported the mean and standard deviation of ROC-AUC across all testing tasks. The evaluation was performed for our model and all baseline methods using support set sizes of 10 and 20, corresponding to 5-shot and 10-shot settings, respectively. Considering that 1-shot learning is impractical in real-world drug discovery scenarios, we excluded 1-shot learning experiments from our study.

\textbf{Baselines.} For a comprehensive comparison, we adopt two types of baselines: (1) methods with molecular encoders trained from scratch, including Siamese \cite{koch2015siamese}, AttnLSTM \cite{lin2016attention}, CHEF \cite{DBLP:journals/corr/abs-2010-06498}, ProtoNet \cite{snell2017prototypical}, MAML \cite{finn2017model}, TPN \cite{liu2018learning}, BOIL \cite{oh2021boilrepresentationchangefewshot}, EGNN \cite{kim2019edge}, IterRefLSTM \cite{altae-tran2017low} and UniMatch \cite{li2025unimatchuniversalmatchingatom}; and (2) methods utlizing pre-trained encoders, including PAR \cite{liu2019learning}, MetaGAT \cite{lv2024meta} and APN \cite{Hou2024}. More details about these baselines are showed in Appendix \ref{baseline}.

\subsection{Main Results}
We evaluate the performance of AdaptMol against all baseline models. The detailed evaluation results are presented in Table \ref{tab:performance_all}. Our observations reveal that AdaptMol consistently achieved state-of-the-art performance across different datasets. In the 5-shot tasks, AdaptMol outperformed the best baseline models with an average improvement of 4.18\% . Moreover, in the 10-shot tasks, AdaptMol demonstrated average improvements of 2.44\% compared to the best-performing baselines. These results substantiate the effectiveness of our model. 

\subsection{Evaluation of Generalization Capability}
To assess the generalization capability of AdaptMol, we constructed the TDC dataset using all classification tasks available on the TDC platform \cite{huang2021therapeuticsdatacommonsmachine}. The detailed content of the TDC dataset is introduced in the Appendix \ref{tdc}. As the training and test sets originate from distinct domains, this task serves as a benchmark for cross-domain generalization. Table \ref{tab:method_comparison} presents the performance of AdaptMol and the state-of-the-art baseline model APN and MetaGAT on 10-shot classification tasks conducted on the TDC dataset.

\begin{table}[!ht]
    \caption{ROC-AUC scores with standard deviations (\%) of all compared methods on TDC dataset. The best results are highlighted in bold font.
    \label{tab:method_comparison}}%
    \centering
    \resizebox{\textwidth}{!}{ 
    \begin{tabular}{lcccccc}
        \toprule
        \multirow{2}{*}{Moldel} & \multicolumn{3}{c}{5-shot} & \multicolumn{3}{c}{10-shot} \\
        \cmidrule(lr){2-4} \cmidrule(lr){5-7}
        & ROC-AUC & F1-Score & PR-AUC & ROC-AUC & F1-Score & PR-AUC \\
        \midrule
            APN & 61.29 (1.23) & 59.41 (1.35) & 59.67 (1.23) & 63.13 (1.28) & 62.55 (1.37) & 62.32 (0.88)\\
            MetaGAT & 62.78 (1.57) & 63.40 (3.89) & 62.61 (0.22) & 64.26 (2.57) & 60.26 (2.40) & 64.66 (2.62) \\
            AdaptMol & \textbf{66.12 (0.76)} & \textbf{63.45 (0.27)} & \textbf{65.54 (0.76)} & \textbf{69.08 (1.00)} & \textbf{64.39 (0.57)} & \textbf{68.81 (1.03)} \\
        \bottomrule
    \end{tabular}
    }
\end{table}

\subsection{Interpretation Case Study}

\begin{figure*}[h]
    \centering
    \includegraphics[width=\textwidth]{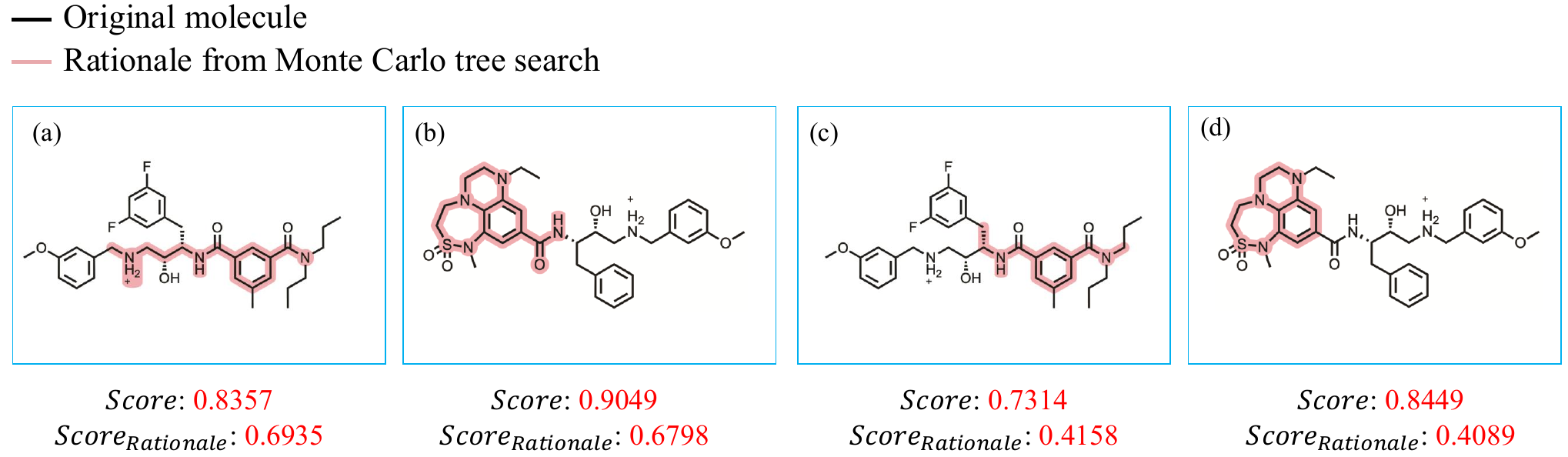} 
    \caption{Using AdaptMol as the scorer for (a) and (b), and single GIN as the scorer for (c) and (d), Monte Carlo Tree Search (MCTS) was employed to extract molecular rationales, which were highlighted within the original molecules. The associated scores for these rationales are presented beneath the figure.}
    \label{fig:case_study}
\end{figure*}

To illustrate the interpretability of our AdaptMol model, we selected representative molecules from the BACE inhibitor dataset and analyzed two examples. Using AdaptMol and a single GIN model as scorers, we applied Monte Carlo Tree Search (MCTS) to identify key substructures (rationales) driving BACE inhibitor activity and their corresponding prediction scores. Figure \ref{fig:case_study} highlights critical substructures, such as amide bonds and secondary amine groups, due to their essential roles in molecular activity. The carbonyl group in the amide bond acts as a hydrogen bond acceptor, enabling interactions with hydrogen-donating residues of the target protein. This interaction, combined with the structural rigidity of the amide moiety, helps maintain a conformation suited to the BACE active site. Additionally, the positively charged secondary amine enhances binding affinity through electrostatic interactions with the anionic region of BACE. Unlike models limited to single molecular graph representations, which often overlook spatial conformations and adaptive behaviors, the AdaptMol model leverages multimodal features to capture complex interactions—such as hydrophobic contacts, hydrogen bonding, and electrostatic forces. This enables a more holistic and accurate understanding of molecular properties and their functional relevance.

\begin{table}[t]
    \centering
    \caption{Results of the ablation study on the multi-level Attention mechanism in DMA. The ROC-AUC scores (\%) with standard deviations for performance on the Tox21 dataset are reported.}
    \label{tab:image_rendering}
    \resizebox{\textwidth}{!}{ 
    \begin{tabular}{ccccccccc}
        \toprule
        \multirow{2}{*}{Local} & \multirow{2}{*}{Global} & \multirow{2}{*}{Adaptive} & \multicolumn{2}{c}{Tox21} & \multicolumn{2}{c}{SIDER} & \multicolumn{2}{c}{MUV} \\
        \cmidrule(lr){4-5} \cmidrule(lr){6-7} \cmidrule(lr){8-9}
         & & & 5-shot & 10-shot & 5-shot & 10-shot & 5-shot & 10-shot\\
        \midrule
        \ding{55} & \ding{55} & - & 80.36 (0.98) & 81.56 (0.85) & 77.56 (0.80) & 79.52 (0.53) & 57.32 (1.57)& 59.23 (0.66)\\
        \ding{51} & \ding{55} & - &  80.65 (0.46) & 82.68 (0.61) & 78.18 (0.42) & 79.83 (0.83) & 70.73 (0.81) & 74.87 (0.55)\\
        \ding{55} & \ding{51} & - & 81.02 (0.39) & 82.28 (0.93) & 78.37 (0.43) & 80.01 (0.59) & 70.03 (0.51) & 76.55 (0.43)\\
        \ding{51} & \ding{51} & \ding{55} & 82.11 (0.56) & 82.33 (0.73) & 78.66 (0.41) & 80.59 (0.23) & 70.65 (1.22) & 76.76 (0.87)\\
        \ding{51} & \ding{51} & \ding{51} & \textbf{83.79 (0.21)} & \textbf{84.93 (0.27)} & \textbf{79.60 (0.61)} & \textbf{81.59 (0.33)} & \textbf{71.65 (0.56)} & \textbf{77.16(0.54)}\\
        \bottomrule
    \end{tabular}
    } 
\end{table}

\subsection{Ablation Study}
Table \ref{tab:image_rendering} presents the results of the ablation study on the multi-level fusion mechanism in AMA. It can be observed that employing either local-level or global-level fusion for integration can partially address the limitation of GNNs in capturing global information. Nevertheless, directly applying multi-level fusion yields marginal performance improvement, as it often introduces redundant information that hinders effective representation learning. In contrast, leveraging adaptive multi-level fusion significantly enhances the performance of GNNs. Specifically, it improves the ROC-AUC by 3.43\% in the 5-shot task and by 3.37\% in the 10-shot task. We also conducted ablation studies on various GNN architectures, and more details see Appendix \ref{abgnn}.

\section{Conclusion}

In this study, we present AdaptMol to address the prevailing challenges associated with few-shot molecular property prediction (MPP). AdaptMol effectively captures multimodal molecular features and incorporates an adaptive fusion mechanism to elucidate the relationship between graph structures and their associated features. This approach achieves state-of-the-art performance across a wide range of molecular property prediction benchmarks. Additionally, we integrated an interpretability-driven methods to identify rationales that determine the key properties of molecules. This approach enhances the transparency of the model's reasoning process, elucidates the importance of dynamically fused multimodaity in augmenting the model's representational capabilities, and offers novel insights for future drug discovery leveraging molecular multimodal representations.

\medskip

\small
\bibliography{main}


\newpage
\appendix

\section*{Appendices}
\section{Algorithm}\label{algori}
In order to clearly describe the training process of AdaptMol framework, we show the process in Algorithm \ref{alg:algorithm}.
\begin{algorithm}[ht]
    \caption{Meta-training procedure for AdaptMol.}
    \label{alg:algorithm}
    \textbf{Require}: A set of tasks for predicting molecular properties $T$\\
    \textbf{Ensure}: AdaptMol parameters $\theta$\\
    Randomly initialize $\theta$
    \begin{algorithmic}[1] 
        \WHILE{not done}
        \STATE Sample a batch of tasks $T_{\tau} \sim T$
        \FORALL{$T_{\tau}$}
        \STATE Sample support set $S_{\tau}$ and query set $Q_{\tau}$ from $T_{\tau}$
        \STATE Obtain sequence features $a_{\tau,i}$ and all atom embedding $G_{\tau,i}$ for each molecular $x_{\tau,i}$
        \STATE Refine $G_{\tau,i}$ to get refined molecular representation by Equation \ref{equa1} - \ref{equa9}
        \STATE Calculate prototype for every class by Equation \ref{equa10}
        \ENDFOR
        \STATE Update $\theta$ by Equation \ref{equa11}
        \ENDWHILE
    \end{algorithmic}
\end{algorithm}

\section{Details of datasets}
\subsection{Details of MoleculeNet dataset}\label{dataset}
To assess the effectiveness and interpretability of our algorithm in molecular property prediction, we conducted experiments on four MoleculeNet datasets \cite{wu2018moleculenet}, detailed as follows:
\begin{itemize}
    \item \textbf{Tox21: }This dataset contains toxicity information of 7831 molecules in 12 assays (each assay
corresponds to a specific target), among which 9 assays are
split for training and 3 assays are split for testing.
    \item \textbf{SIDER: }This dataset records the side effects information of 1427 compounds in 27 classes, among which 21 classes are split for training and 6 classes are split for testing.
    \item \textbf{MUV: } This dataset is designed to provide a challenging benchmark for virtual screening methods. It consists of 93127 compounds in 17 assays, among which 12 assays are split for training and 5 assays are split for testing.
    \item \textbf{BACE: } This dataset provides quantitative (IC50) and qualitative (binary) binding results for a set of inhibitors of human $\beta-secretase 1$ (BACE-1). It includes 1,522 compounds, offering a platform for evaluating regression and classification models in drug discovery contexts. 
\end{itemize}

\subsection{Details of TDC dataset}\label{tdc}

\begin{table*}[!ht]
\caption{The detail information of TDC datasets.}
\label{tab:tdc_datasets}
\vspace{6pt}
\begin{tabular*}{\textwidth}{@{\extracolsep{\fill}}lcccccc@{\extracolsep{\fill}}}
\toprule%
No. & Dataset & Sample & Type \\
\midrule
        1 & hia\_hou & 578 & Absorption \\
        2 & pgp\_broccatelli & 1218 & \\
        3 & bioavailability\_ma & 640 & \\
        4 & bbb\_martins & 2030 & Distribution \\
        5 & cyp2c9\_substrate\_carbonmangels & 669 & Metabolism \\
        6 & cyp2d6\_substrate\_carbonmangels & 667 & \\
        7 & cyp3a4\_substrate carbonmangels & 670 & \\
        8 & herg & 655 & Toxicity \\
        9 & ames & 7278 & \\
        10 & dili & 475 & \\
\bottomrule
\end{tabular*}
\end{table*}

The TDC dataset is meticulously designed to assess the generalization capabilities of models across critical pharmacological endpoints \cite{huang2021therapeuticsdatacommonsmachine}. It includes three absorption datasets, one distribution dataset, and three metabolism datasets for training, along with three toxicity datasets designated for testing. The detailed information is presented in Table \ref{tab:tdc_datasets}.

\section{Details of baselines}\label{baseline}
\paragraph{Methods with Molecular Encoders Trained from Scratch:}
\begin{itemize}
  \item \textbf{Siamese}~\cite{koch2015siamese}: Employs a dual-network architecture to assess similarity between molecular pairs, facilitating pairwise comparison tasks.
  
  \item \textbf{AttnLSTM}~\cite{lin2016attention}: Integrates attention mechanisms with Long Short-Term Memory networks to capture relevant substructures in molecular sequences for property prediction.
  
  \item \textbf{CHEF}~\cite{DBLP:journals/corr/abs-2010-06498}: Utilizes handcrafted features combined with ensemble learning techniques to predict molecular properties from structural information.
  
  \item \textbf{ProtoNet}~\cite{snell2017prototypical}: Learns a metric space where classification is performed by computing distances to prototype representations of each class, enabling few-shot learning.
  
  \item \textbf{MAML}~\cite{finn2017model}: Applies Model-Agnostic Meta-Learning to acquire initial parameters that can be rapidly adapted to new tasks with limited data through gradient updates.
  
  \item \textbf{TPN}~\cite{liu2018learning}: Constructs a task-specific graph to propagate labels from labeled to unlabeled instances, leveraging the manifold structure of the data for transductive inference.
  
  \item \textbf{BOIL}~\cite{oh2021boilrepresentationchangefewshot}: Focuses on representation learning by emphasizing the importance of feature extraction over classifier adaptation in few-shot scenarios.
  
  \item \textbf{EGNN}~\cite{kim2019edge}: Predicts edge labels within a graph constructed from input samples to explicitly model intra-cluster similarity and inter-cluster dissimilarity.
  
  \item \textbf{IterRefLSTM}~\cite{altae-tran2017low}: Adapts Matching Networks by incorporating iterative refinement through LSTM-based attention mechanisms for molecular property prediction.
  
  \item \textbf{UniMatch}~\cite{li2025unimatchuniversalmatchingatom}: Implements a unified matching framework that aligns query and support instances in a shared embedding space to facilitate few-shot learning tasks.
\end{itemize}

\paragraph{Methods Utilizing Pre-trained Encoders:}
\begin{itemize}
  \item \textbf{PAR}~\cite{liu2019learning}: Employs class prototypes to update input representations and designs label propagation mechanisms within a relational graph to transform generic molecular embeddings into property-aware spaces.
  
  \item \textbf{MetaGAT}~\cite{lv2024meta}: Integrates meta-learning with Graph Attention Networks to capture task-specific information, enhancing the adaptability of molecular representations across diverse property prediction tasks.
  
  \item \textbf{APN}~\cite{Hou2024}: Leverages attention-based prototype networks to refine molecular embeddings, facilitating effective few-shot learning by focusing on relevant substructures associated with specific properties.
\end{itemize}

\section{Implementation details}
We implement the AdaptMol architecture primarily using PyTorch and employ the Adam optimizer \cite{kingma2014adam} for training. The learning rate is set within the range of 0.0005 to 0.05 to facilitate effective gradient descent optimization. Regarding the crucial hyperparameter settings of dynamic modal weights in Equation \ref{betas} and \ref{betag}, we set the scaling factor $k = 2$, while $\beta_\text{min} = 0.9$ and $\beta_\text{max} = 1.1$. The AdaptMol architecture was trained on a NVIDIA GeForce RTX 2080 Ti GPU, paired with an Intel(R) Xeon(R) CPU E5-2678 v3 @ 2.50GHz, running on the Ubuntu 18.04 platform. During training, 2000 episodes were generated under a 2-way 10-shot setting. For the classification task, cross-entropy loss was employed as the objective function, and an early stopping strategy was implemented with a patience level of 100 to prevent overfitting. During the testing phase, consistent with the approach outlined in \cite{lv2024meta}, we randomly sampled support sets of size 10 or 20 and query sets of size 32 from the test tasks. To ensure robustness and minimize the influence of randomness, each test task was evaluated over 10 independent runs with different random seeds. The final performance of our model was determined by averaging the results across all runs.

\begin{figure*}[!ht]
    \centering
    \includegraphics[width=\textwidth]{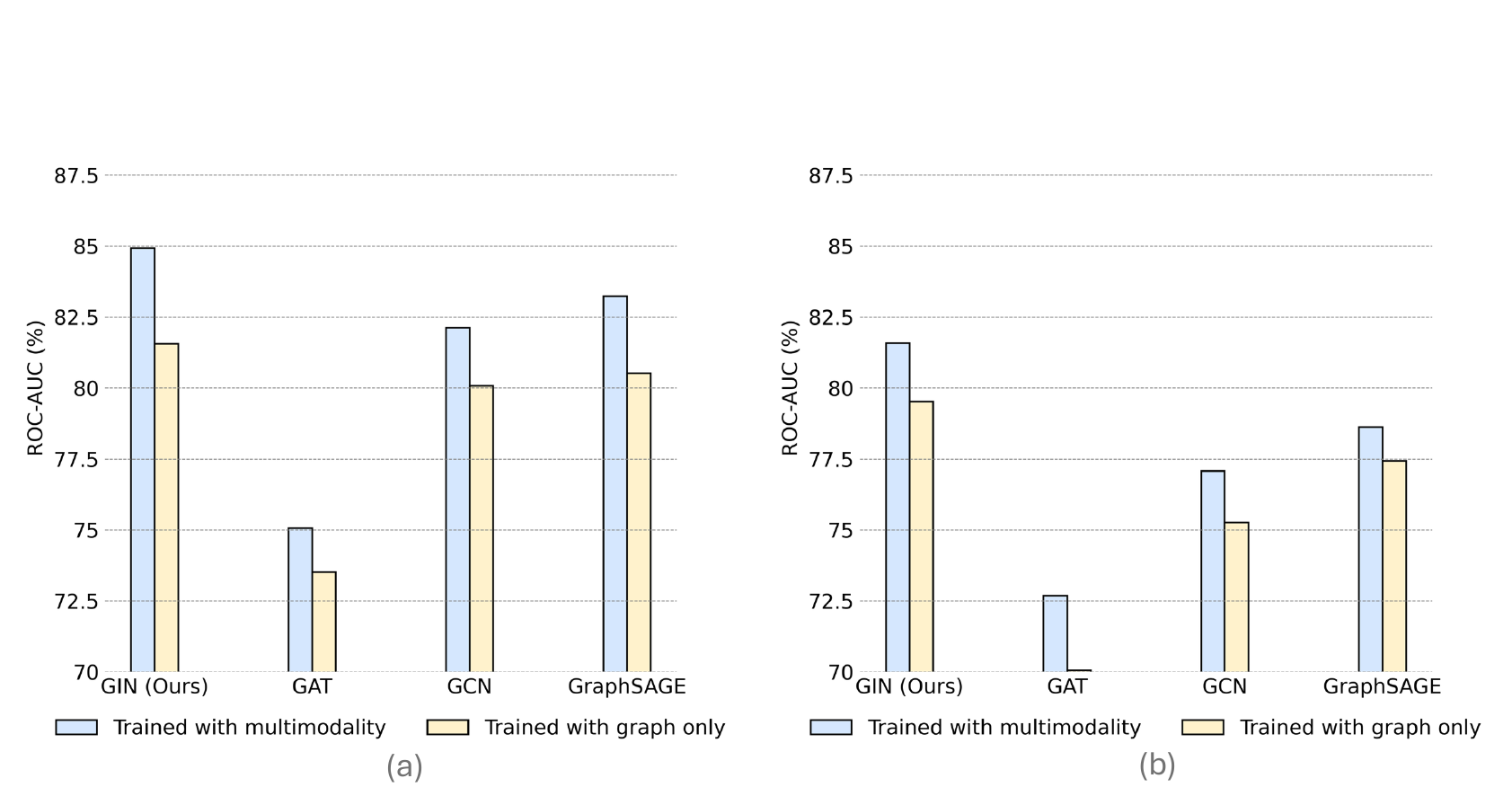} 
    \caption{(a) ROC-AUC performance from Tox21 datasets. (b) ROC-AUC performance from SIDER datasets.}
    \label{fig:gnn}
\end{figure*}

\begin{figure*}[!ht]
    \centering
    \includegraphics[width=\textwidth]{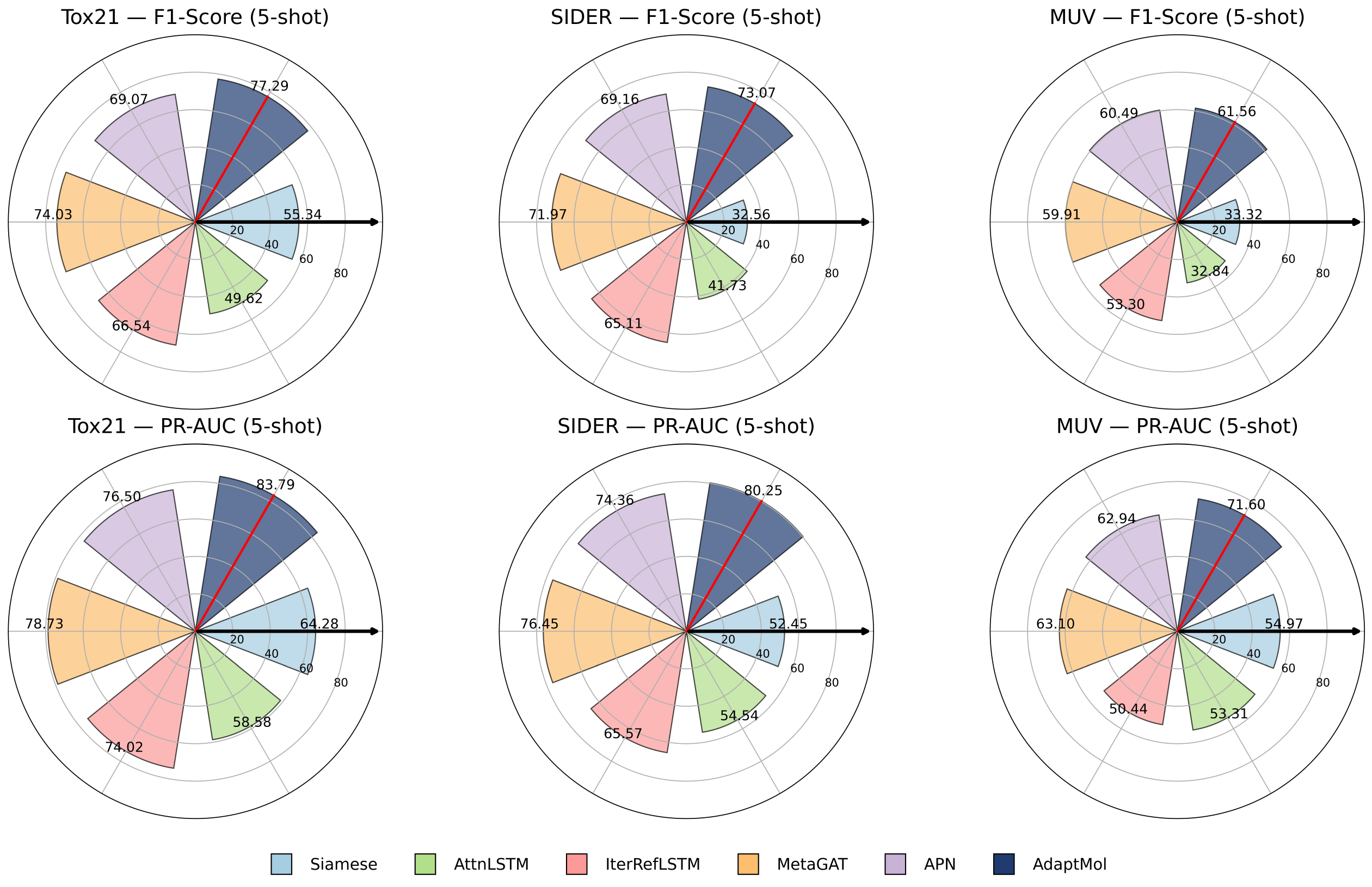} 
    \caption{The additional performance of all compared methods on three tasks with a support set of size 10 on the MoleculeNet benchmark. Each colored sector corresponds to a specific method, where the length of the sector reflects its performance based on F1-score and PRAUC (\%). Starting from the horizontal right-pointing arrow, the methods are listed in the legend in a counterclockwise direction. Our \textbf{AdaptMol} corresponds to the last dark blue sector.}
    \label{add}
\end{figure*}

\section{Additional experiments results}
\subsection{Ablation Study on different GNN architecture}\label{abgnn}
We have introduced the Graph Encoder employed in our model, which can be substituted with alternative graph-based molecular encoders. To demonstrate the superior molecular representation capability of our model, we evaluated it using three additional molecular graph encoders: GCN, GAT, and GraphSAGE. Figure \ref{fig:gnn} (a) and Figure \ref{fig:gnn} (b) present the ROC-AUC performance achieved on 10-shot tasks from Tox21 and SIDER datasets, respectively.

\subsection{Additional metrics on MoleculeNet}
To provide a comprehensive comparison with our model, we conduct a series of additional experiments on the MoleculeNet benchmark and report both the F1-score and PRAUC. Specifically, the F1-score offers a holistic evaluation of classification performance by balancing precision and recall, while PRAUC is particularly suitable for tasks with highly imbalanced distributions, such as MUV. We compare our model against five representative baseline methods, including Siamese, AttnLSTM, IterRefLSTM, MetaGAT, and APN. Figure \ref{add} illustrates the detailed information. The results indicate that our model consistently achieves state-of-the-art performance on two additional critical classification metrics. Across three datasets, it surpasses the best-performing baseline by an average of 1.81\% in F1-score and 5.79\% in PRAUC, highlighting its strong molecular representation ability. Furthermore, the model demonstrates remarkable stability on the imbalanced MUV dataset.

\section{Limitation and future directions}
\textbf{Limitation: }Despite achieving state-of-the-art performance on most few-shot tasks, the proposed adaptive fusion mechanism is still relatively simplistic. In particular, for molecules with simple structures, it may lead to information redundancy, thereby limiting the effectiveness of the molecular representations learned by the model. 

\textbf{Future Work: }In the future, we will seek to develop more expressive and flexible fusion architectures to enhance the model’s representational capacity. For instance, we plan to investigate fine-grained fusion schemes that operate at different structural levels (e.g., atom, bond, and substructure) and adaptively weight their contributions based on molecular context. Such schemes could leverage hierarchical attention mechanisms or learnable gating networks to capture salient features more effectively and reduce redundancy. Moreover, we aim to incorporate automated optimization of the fusion strategy— for example, by employing neural architecture search or meta-learning techniques—so that the most appropriate fusion parameters are discovered in a data-driven fashion and adjusted for each molecule. Overall, these directions aim to push the boundaries of molecular representation learning by developing fusion strategies that are both more powerful and more broadly applicable.
\end{document}